\documentclass[11pt]{article}

\usepackage[preprint]{acl}

\usepackage{times}
\usepackage{latexsym}
\usepackage[T1]{fontenc}
\usepackage[utf8]{inputenc}
\usepackage{microtype}
\usepackage{inconsolata}

\usepackage{graphicx}
\usepackage{subcaption}
\usepackage{float}
\usepackage{booktabs}
\usepackage{multirow}
\usepackage{amsmath}

\newcommand{\bfcl}{BFCL~v4}
\newcommand{\codemode}{programmatic tool calling}
\newcommand{\Codemode}{Programmatic tool calling}
\newcommand{\jsonmode}{JSON tool calling}
\newcommand{\Jsonmode}{JSON Tool Calling}
\newcommand{\execpy}{\texttt{execute}}

\newcommand{\nmodels}{14}

\title{The Bitter Lesson of Tool Calling}

\author{
 \textbf{Ishan Patel}, 
 \textbf{Sahil Sen}, 
 \textbf{Elias Lumer}, 
 \textbf{and Vamse Kumar Subbiah} \\
 \
 \textit{Commercial Technology and Innovation Office} \\\textit{PricewaterhouseCoopers, U.S.A}}

\date{}

\begin{document}

\maketitle

\begin{abstract}
Tool use transforms LLMs into agents that act beyond their training data, and for
code-capable models, programmatic tool calling extends this further by replacing rigid JSON
calls with scripts that chain and parallelize naturally.
However, a systematic evaluation of tools as code on an established benchmark across
current and prior model generations under real-world task conditions has not been conducted.
In this work, we empirically compare programmatic tool calling (PTC) to native JSON tool calling
across \nmodels{} language models on \bfcl{}.
In the programmatic tool calling paradigm, tools are exposed as typed Python stubs that the model invokes
through code, with execution and results handled in a single agent turn.
Programmatic tool calling matches or exceeds native \jsonmode{} in 11 of 14 models on \bfcl{},
with the GPT-5.6 family achieving a 10.6\% improvement over the \jsonmode{} baseline.
Further, it matches or outperforms baseline in 13 of 14 models under parallel fan-out,
and holds stable under context rot conditions where baseline degrades 2.3\% on average.
Our results demonstrate that programmatic tool calling is a viable and robust alternative to
\jsonmode{}, with performance tracking model capability across release generations.
\end{abstract}

\section{Introduction}

Large language models (LLMs) now act as tool-calling agents, invoking external
services through APIs that require the model to emit a structured JSON object at
each function call~\citep{anthropicagents2024, deterministic2026}.
For models that can already write executable code, this format is a design
choice, not a necessity.

Prior work has established a theoretical case for replacing JSON tool calls with
executable code.
CodeAct~\citep{wang2024codeact} showed that code actions achieve up to 20\%
higher task success with 30\% fewer interaction turns on multi-tool tasks, with
gains concentrated in parallel and compositional scenarios.
A more recent study found that code-only output restrictions change pass rates by
an absolute change of fewer than 3\% on coding-agent tasks~\citep{kddcoding2026}.
Function calling is a harder test case. It requires precise argument serialization,
multi-step chaining, and fan-out across heterogeneous APIs, constraints absent
from coding-agent evaluations and exactly where paradigm choice has the largest
claimed effect.
Whether programmatic tool calling matches \jsonmode{} under these conditions,
across model families and adversarial context loads, remains untested.

To address this gap, we evaluate \textit{\codemode{}}
against \jsonmode{} on a representative 309-entry subset of \bfcl{}
spanning eight task categories, using \nmodels{} models released between November~2024
and July~2026.
In \codemode{}, the model writes a Python script using typed Python stubs compiled
from the benchmark's function schemas. The agent loop executes it in a shell
subprocess, producing tool-call results without additional inference turns.
We run three ablation studies targeting sequential chaining, parallel fan-out, and
context rot, the task structures where the limitations of \jsonmode{} have been
most commonly claimed.

We find that 11 of \nmodels{} models match or exceed baseline accuracy
under \codemode{} on \bfcl{}, with results dividing along model generation lines.
All five Anthropic models and the three newest GPT generations match or exceed
baseline accuracy, while three older GPT models do not.
On context flooding, \codemode{} sees an absolute improvement of 5.5\% on
average, compared to an absolute 32\% degradation for the
filesystem-discovery comparison approach.

Our contributions are:
\begin{itemize}
  \item \Codemode{} viability divides along model generation lines rather than
  model family. All five Anthropic models and the three newest GPT generations
  match or exceed \jsonmode{} accuracy on \bfcl{}, while three older GPT models
  do not.
  \item \Codemode{}'s accuracy advantage on sequential tasks scales with chain
  length, reaching an 18.8\% absolute gap over \jsonmode{} at lengths $\geq 12$,
  an effect absent at short chains and driven by the extra inference turn
  \jsonmode{} incurs per link.
  \item \Jsonmode{} drops tool calls entirely above a model-specific fan-out
  threshold ($N=70$--$72$ for Claude Sonnet~5). \Codemode{} maintains 100\%
  enumeration accuracy at $N=100$, exposing a hard structural limit in the
  native paradigm.
  \item Under context flooding, \codemode{} holds stable while \jsonmode{}
  degrades 2.3\% on average and the filesystem-discovery approach degrades
  32\%, demonstrating robustness to adversarial context load.
\end{itemize}

\section{Related Work}

Two lines of work motivate replacing JSON tool calls with executable code.
The first is code-action agents. CodeAct~\citep{wang2024codeact} showed that
code actions outperform JSON alternatives on multi-tool tasks, with gains
concentrated in parallel and compositional scenarios.
Recursive Agent Harnesses~\citep{lumer2026recursive} extends the same
code-execution primitive to parallel subagent spawning, showing that executing
code in a subprocess bypasses the per-turn tool-call cap that constrains native
JSON function calling at high fan-out.
Practitioner frameworks extended this finding into production, treating code
composition as the default interface~\citep{smolagents2024,anthropicagents2024,primeagent2026}.
Production memory systems such as Chronos~\citep{sen2026chronos} deploy iterative
JSON tool-calling loops as their core agent mechanism, confirming that JSON tool
calling remains the dominant interface in deployed agentic systems.
Infrastructure providers are adopting the same pattern. Cloudflare's Agents platform
ships code execution as its native tool-use interface (currently experimental)~\citep{cloudflarecodemode2026}.
A recent survey argues that code is the ideal substrate for agentic reasoning
because it is executable, inspectable, and stateful~\citep{codeharness2026}.
The Deterministic Horizon~\citep{deterministic2026} provides theoretical grounding.
Tool delegation is necessary precisely where chain-of-thought reasoning fails to
produce verifiable answers, the same regime where tool-calling accuracy is
measured.
However, none of these works evaluate code-execution against \jsonmode{}
on a standardized benchmark across multiple model families.

Several benchmarks measure LLM tool-calling accuracy, including
API-Bank~\citep{chen2024apibank}, T-Eval~\citep{chen2024teval},
API-BLEND~\citep{acharya2024apiblend}, UltraTool~\citep{du2024ultratool},
CONFETTI~\citep{hosseini2025confetti}, and ToolHop~\citep{wang2025toolhop}.
Each evaluates how accurately models call the right function with the right
arguments, but none compares paradigms.
\bfcl{} covers eight task categories with a deterministic scorer. A recent audit~\citep{bfclaudit2026} found 20\%
evaluator-human misalignment in its LLM-judge mode, which our evaluation
sidesteps by scoring against stub outputs directly.
Work on agentic tool-calling training efficiency~\citep{multiturn2026} and context
inflation~\citep{hypertool2026} identifies format and prompt engineering as
dominant sources of variance in tool-calling benchmarks, motivating the
controlled three-way paradigm comparison we conduct.
Unlike prior work, we evaluate \codemode{} against \jsonmode{} on
\bfcl{} across \nmodels{} models spanning 20~months of releases, with three
ablation studies targeting chaining, parallel fan-out, and context rot.

\section{Method}

We study whether programmatic tool calling can replace \jsonmode{}
for LLM tool use without sacrificing accuracy.
We evaluate two paradigms on \bfcl{}, \textit{JSON tool calling} and \textit{\codemode{}}.
In each paradigm, the model receives the same task description and must invoke
the correct functions with the correct arguments. Only the mechanism for
expressing and executing tool calls differs.
In the context rot ablation we additionally include a filesystem-based condition
as an internal reference point. It is not an industry practice and is referenced
only where it illuminates the primary comparison.

\begin{figure*}[t]
  \centering
  \includegraphics[width=\textwidth]{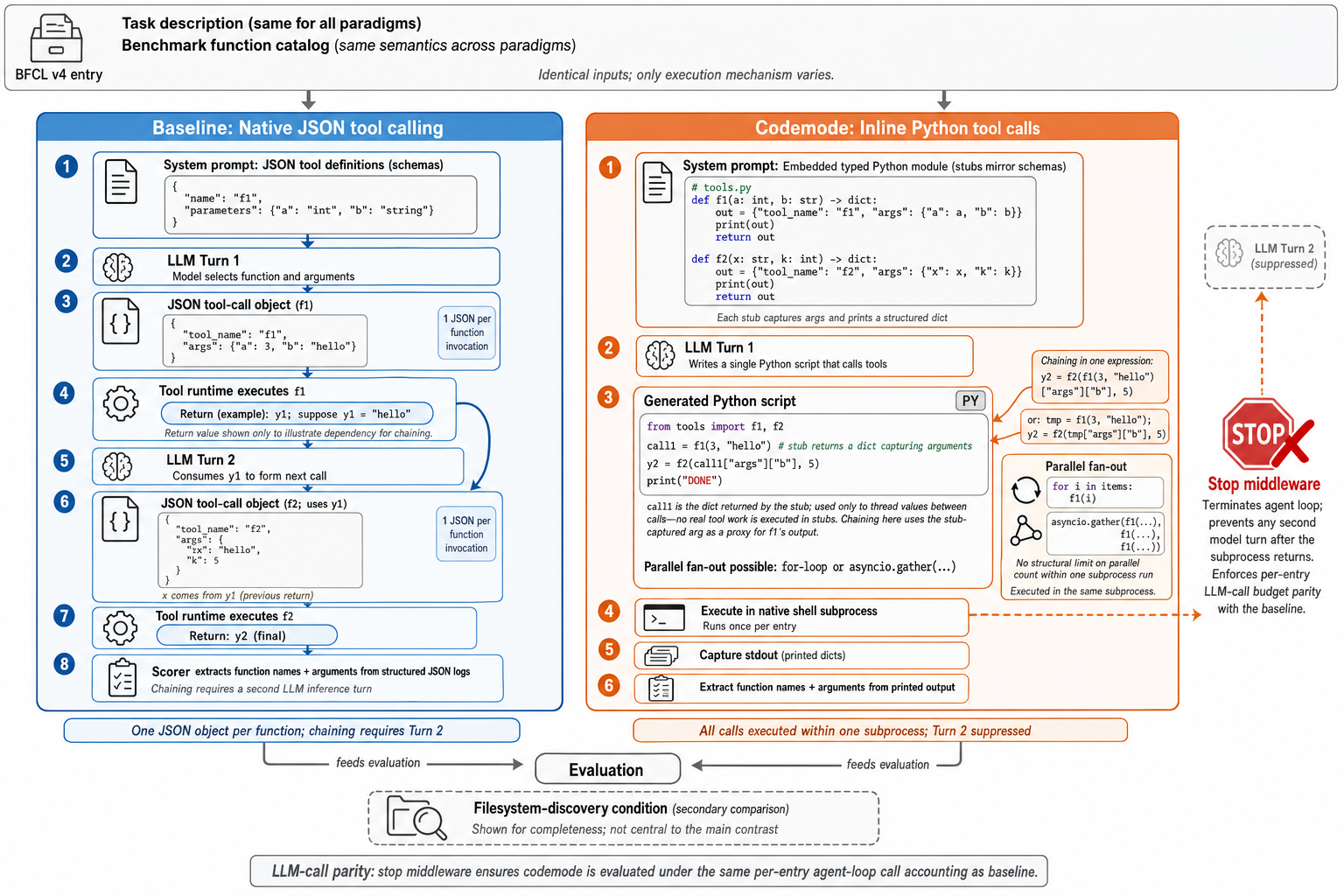}
  \caption{Overview of the two primary paradigms evaluated. In \textit{JSON tool calling}, the
  model emits JSON tool-call objects via the API. In \textit{\codemode{}},
  it writes a Python script using typed stubs; the agent loop executes it in a
  subprocess. A filesystem-discovery condition is included as a secondary reference point.}
  \label{fig:architecture}
\end{figure*}

\subsection{Task Definition}

Each benchmark entry specifies a natural-language user query $q$, a set of
available functions $\mathcal{F} = \{f_1, \ldots, f_k\}$ each with a typed
signature, and a ground-truth set of function calls
$\mathcal{C}^* = \{(f_i, \mathbf{a}_i)\}$ where $\mathbf{a}_i$ is the argument
mapping for call $i$.
A model is correct on an entry if and only if its output produces a set of calls
that matches $\mathcal{C}^*$ under the benchmark's normalized string
comparison, which strips punctuation and lowercases all values before matching.
This definition is identical across all three paradigms. Differences
arise only in how the model is prompted and how its output is parsed.

\subsection{Paradigms}

\paragraph{JSON Tool Calling.}
The model receives a system prompt containing the function schemas formatted as
JSON tool definitions and an API call that invokes the tool-calling endpoint.
The model outputs a structured JSON object at each function call.
This is the standard deployment pattern for tool-augmented LLMs and serves as
our reference condition.

\paragraph{\Codemode{}.}
The model receives a system prompt that embeds the source of a typed Python
module whose functions correspond one-to-one with the benchmark's function
schemas.
Each stub function captures its arguments and returns them as a structured dict.
No external service is contacted.
The model writes a Python script that imports this module and calls the
appropriate functions.
The agent loop executes the script in a native shell subprocess. The subprocess
captures stdout, and the scorer extracts function names and argument values
from the printed output.
No additional inference turns occur after the subprocess returns. A stop
middleware intercepts the next model call and terminates the agent loop.
This design ensures that \codemode{} and JSON tool calling consume the same number of LLM
calls per entry, making accuracy comparisons straightforward.

Consider a task that asks the model to retrieve the circumference of a circle
with radius 7 and the area of a square with side 5.
In JSON tool calling, the model emits two sequential JSON tool-call objects.
In \codemode{}, the model writes:

{\small\begin{verbatim}
execute(command="python3 -c '
import json
from stubs import (
    circumference, area_square)
print(json.dumps(circumference(radius=7)))
print(json.dumps(area_square(side=5)))
'")
\end{verbatim}
}

The shell subprocess captures both lines of stdout, and the scorer
matches each printed call against the ground truth.
\Codemode{} resolves both calls in a single
Python expression evaluated in one subprocess, while JSON tool calling requires two
separate model outputs.

\subsection{Benchmark and Evaluation}

\bfcl{}~\citep{bfclaudit2026} provides 309 representative entries drawn from eight task categories:
\textit{simple\_python}, \textit{multiple}, \textit{parallel},
\textit{parallel\_multiple}, \textit{live\_simple}, \textit{live\_multiple},
\textit{live\_parallel}, and \textit{live\_parallel\_multiple}.
Entries were sampled proportionally to mirror the category distribution of the full \bfcl{} benchmark,
with per-category minimums applied to ensure sufficient entries in smaller categories.
The complete entry ID list is released with the evaluation harness.
Simple and multiple categories test single-call and multi-call accuracy on
curated queries. Live categories use queries sampled from real user interactions.
Parallel categories require the model to issue two or more function calls whose
arguments are independent of each other.

All three paradigms use a deterministic scorer that normalizes predicted and
ground-truth argument values before comparison.
We report accuracy as the fraction of entries where every required function call
is both present and correctly parameterized.
When the shell subprocess raises a syntax or runtime error, the entry is scored
as zero correct calls. No entries are skipped or excluded from reported totals.

\subsection{Ablation Design}

The three ablation studies target task structures where paradigm choice has been claimed
to matter.
Each ablation subset was constructed by selecting entries from \bfcl{} that
represent the target task structure without duplicating main-evaluation entries.
The complete entry ID lists are released with the evaluation harness.
The chaining subset ($n=52$) was optimized to cover the full range of chain
lengths ($n_{\text{chain}}=2$--$20$), with the distribution weighted toward
longer chains where JSON tool calling and \codemode{} diverge most ($n_{\text{chain}}\geq 6$:
17~entries; $n_{\text{chain}}\geq 13$: 10~entries).
The parallelism subset ($n=32$) covers fan-out counts of 7 to 48 (7, 9, 11, 13, 15, 20, 30, 48),
drawn from a synthetic corpus. We report only enumeration-type entries ($n=32$ of 48 total),
excluding aggregation-type entries from the accuracy figures.
To locate the fan-out threshold at which Anthropic frontier models begin dropping calls under
JSON tool calling, we extend the corpus with probe entries at $N \in \{60, 70, 72, 75, 100\}$
and evaluate Claude Sonnet~5 baseline and \codemode{} on those entries.
The context rot subset ($n=31$ per condition) uses entries from
\textit{live\_multiple}, \textit{live\_parallel}, and
\textit{live\_parallel\_multiple} categories where function schemas are most
easily contaminated by domain-unrelated decoys.

\paragraph{Chaining.}
Sequential multi-hop function calls, where the output of $f_1$ must be computed
and passed as an argument to $f_2$.
In JSON tool calling, this requires two model turns. The model calls $f_1$, receives the
return value, then calls $f_2$.
In \codemode{}, both calls appear in a single script. The model computes the
intermediate value using parametric knowledge and passes it directly.
The chaining ablation contains 52 entries with chain lengths of 2 to 20 calls.

\paragraph{Parallelism.}
Independent fan-out function calls, where $N$ calls must be issued in a single
step.
JSON tool calling issues them as parallel tool-call objects. \Codemode{} writes an
\texttt{asyncio.gather} block or a sequential loop.
The parallelism ablation uses 32 enumeration-type entries with fan-out counts ranging from 7 to 48,
plus targeted probe entries at $N \in \{60, 70, 72, 75, 100\}$ for Claude Sonnet~5 to locate
the fan-out threshold at which JSON tool calling begins dropping calls.
We separately report \textit{enumeration accuracy} (did the model issue all $N$
required calls?) and \textit{aggregation accuracy} (did the model produce the
correct aggregate answer?), as \codemode{} can answer aggregation questions from
parametric knowledge without executing every call.

\paragraph{Context rot.}
The flood condition injects decoy function schemas into the context alongside
the entry's relevant schemas, increasing total context size to 128 schemas.
The two conditions are \textit{filtered} (only the functions the query uses)
and \textit{flood} (128 total schemas, including the relevant functions plus
corpus decoys drawn from unrelated domains).
The context rot ablation contains 31 entries per condition.

\subsection{Models}

We evaluate \nmodels{} models spanning two families and 20~months of releases,
listed in Table~\ref{tab:models}.
All models are run at temperature~0.

\begin{table}[t]
\centering
\footnotesize
\setlength{\tabcolsep}{4pt}
\begin{tabular}{ll}
\toprule
\textbf{Name} & \textbf{Model ID} \\
\midrule
\multicolumn{2}{l}{\textit{Anthropic}} \\
Claude Haiku~4.5  & \texttt{claude-haiku-4-5} \\
Claude Sonnet~4.5 & \texttt{claude-sonnet-4-5} \\
Claude Sonnet~4.6 & \texttt{claude-sonnet-4-6} \\
Claude Opus~4.8   & \texttt{claude-opus-4-8} \\
Claude Sonnet~5   & \texttt{claude-sonnet-5} \\
\midrule
\multicolumn{2}{l}{\textit{OpenAI}} \\
GPT-4o        & \texttt{gpt-4o-2024-11-20} \\
GPT-4.1       & \texttt{gpt-4.1-2025-04-14} \\
GPT-5-nano    & \texttt{gpt-5-nano-2025-08-07} \\
GPT-5         & \texttt{gpt-5-2025-08-07} \\
GPT-5.4-mini  & \texttt{gpt-5.4-mini-2026-03-17} \\
GPT-5.4       & \texttt{gpt-5.4-2026-03-05} \\
GPT-5.6-Luna  & \texttt{gpt-5.6-luna} \\
GPT-5.6-Sol   & \texttt{gpt-5.6-sol} \\
GPT-5.6-Terra & \texttt{gpt-5.6-terra} \\
\bottomrule
\end{tabular}
\caption{Models evaluated, covering releases from November~2024 to July~2026.}
\label{tab:models}
\end{table}
All tables report 95\%~Wilson confidence intervals per row.
Because ablation entries are small ($n = 31$--$52$), intervals are wide and
individual model results should be treated as directional rather than conclusive.
Only cross-model aggregate patterns are interpreted as reliable findings.
Pairwise per-model comparisons are not corrected for multiple testing. We make no
claim that any individual model result is statistically significant in isolation.

\section{Experiments}

We evaluate \nmodels{} models on \bfcl{} and on three ablation studies (chaining,
parallelism, and context rot).
All experiments are run at temperature~0.
The results below report accuracy as the fraction of entries where every required
function call is both present and correctly parameterized.

\subsection{BFCL v4 Main Evaluation}

\begin{table}[t]
\centering
\footnotesize
\setlength{\tabcolsep}{3pt}
\begin{tabular}{lrrrr}
\toprule
\textbf{Model} & \multicolumn{2}{c}{\textbf{JSON}} & \multicolumn{2}{c}{\textbf{PTC}} \\
\cmidrule(lr){2-3}\cmidrule(lr){4-5}
 & Acc & $\pm$ & Acc & $\pm$ \\
\midrule
\multicolumn{5}{l}{\textit{Anthropic}} \\
Claude Haiku~4.5  & 81.9 & 4.3 & \textbf{85.8} & 3.9 \\
Claude Sonnet~4.5 & 86.4 & 3.8 & \textbf{87.7} & 3.7 \\
Claude Sonnet~4.6 & 80.9 & 4.4 & \textbf{87.4} & 3.7 \\
Claude Opus~4.8   & \textbf{84.8} & 4.0 & \textbf{84.8} & 4.0 \\
Claude Sonnet~5   & 84.5 & 4.0 & \textbf{86.1} & 3.9 \\
\midrule
\multicolumn{5}{l}{\textit{OpenAI}} \\
GPT-4o            & \textbf{81.9} & 4.3 & 55.0 & 5.5 \\
GPT-4.1           & \textbf{81.9} & 4.3 & 62.1 & 5.4 \\
GPT-5-nano        & 66.7 & 5.2 & \textbf{68.3} & 5.2 \\
GPT-5             & 71.5 & 5.0 & \textbf{76.1} & 4.7 \\
GPT-5.4-mini      & \textbf{79.3} & 4.5 & 55.0 & 5.5 \\
GPT-5.4           & 79.3 & 4.5 & \textbf{81.9} & 4.3 \\
GPT-5.6-Luna      & 76.1 & 4.7 & \textbf{80.3} & 4.4 \\
GPT-5.6-Sol       & 72.2 & 5.0 & \textbf{82.8} & 4.2 \\
GPT-5.6-Terra     & 73.5 & 4.9 & \textbf{84.1} & 4.1 \\
\bottomrule
\end{tabular}
\caption{Accuracy (\%) on our 309-entry \bfcl{} subset. JSON = JSON tool calling;
PTC = programmatic tool calling. $\pm$ columns show 95\%
Wilson CI half-widths. Bold indicates the higher value per row (ties bolded on both).}
\label{tab:bfcl301}
\end{table}

Table~\ref{tab:bfcl301} reports accuracy for JSON tool calling and \codemode{} across all
\nmodels{} models on \bfcl{}.
The GPT-5.6 family achieves the largest \codemode{} gains. GPT-5.6-Sol and
GPT-5.6-Terra each achieve an absolute improvement of 10.6\% over their own JSON tool calling baseline.
Across all \nmodels{} models, 11 match or exceed baseline accuracy under \codemode{}.
Per-category mean accuracies across all 14 models are reported in
Table~\ref{tab:per_category} (Appendix). The aggregate result conceals meaningful
category-level variance.
\Codemode{} underperforms JSON tool calling by an absolute 14.1\% on average in the
\textit{parallel} and \textit{parallel\_multiple} categories, while gaining an absolute 10.0\% in \textit{live\_multiple}.
The parallel-category gap is partially explained by the \texttt{\textbackslash n}
encoding failures of the three weaker OpenAI models, which are most harmful on
multi-call scripts. Excluding those three models narrows the gap.

All five Anthropic models (Claude Haiku~4.5 through Sonnet~5) match or exceed
baseline accuracy under \codemode{}, with deltas ranging from 0.0 to 6.5 percentage
absolute, a pattern that holds across model generations.
Among OpenAI models, the three newest GPT-5.6 variants are all positive (absolute improvements of 4.2\% to 10.6\% over baseline), while three older models (GPT-4o, GPT-4.1, and
GPT-5.4-mini) fall below baseline by an absolute 19.7\% to 26.9\%.
For those three models the failure mode is consistent. The models produce code
with literal \texttt{\textbackslash n} escape sequences in multiline scripts
rather than real newlines, causing the subprocess to fail with a syntax error on
any entry requiring more than a single-line script.
GPT-5-nano, released alongside GPT-5, does not exhibit this failure, suggesting
the fix entered training data between GPT-5.4-mini and GPT-5.

\subsection{Chaining Ablation}

In the chaining ablation ($n = 52$ entries, chain lengths 2--20), \codemode{}
handles sequential multi-hop calls differently from JSON tool calling.
In JSON tool calling, the model issues $f_1$, receives its return value, then issues $f_2$
in a second inference turn.
In \codemode{}, the model writes both calls in a single script and computes the
intermediate value from parametric knowledge before passing it to $f_2$.

Table~\ref{tab:chaining} shows per-model results.
Claude Sonnet~5 achieves the largest \codemode{} gain (80.8\% $\to$ 96.2\%), followed by Claude Opus~4.8 (80.8\% $\to$ 94.2\%), while six
models maintain near-parity (within 5\% absolute).
GPT-4.1 is the sole outlier. Its baseline accuracy of 98.1\% collapses to
40.4\% under \codemode{}, driven by the \texttt{\textbackslash n} encoding failure
described above, which is most harmful in chaining tasks because every
intermediate computation requires a multiline script.

\begin{table}[t]
\centering
\footnotesize
\setlength{\tabcolsep}{3pt}
\begin{tabular}{lrrrr}
\toprule
\textbf{Model} & \multicolumn{2}{c}{\textbf{JSON}} & \multicolumn{2}{c}{\textbf{PTC}} \\
\cmidrule(lr){2-3}\cmidrule(lr){4-5}
 & Acc & $\pm$ & Acc & $\pm$ \\
\midrule
\multicolumn{5}{l}{\textit{Anthropic}} \\
Claude Haiku~4.5  & \textbf{88.5} &  8.8 & 73.1 & 11.7 \\
Claude Sonnet~4.5 & \textbf{90.4} &  8.2 & 88.5 &  8.8 \\
Claude Sonnet~4.6 & 90.4 &  8.2 & \textbf{92.3} &  7.6 \\
Claude Opus~4.8   & 80.8 & 10.6 & \textbf{94.2} &  6.8 \\
Claude Sonnet~5   & 80.8 & 10.6 & \textbf{96.2} &  6.0 \\
\midrule
\multicolumn{5}{l}{\textit{OpenAI}} \\
GPT-4o            & \textbf{92.3} &  7.6 & 86.5 &  9.3 \\
GPT-4.1$^\dagger$ & \textbf{98.1} &  4.9 & 40.4 & 12.9 \\
GPT-5-nano        & 69.2 & 12.2 & \textbf{80.8} & 10.6 \\
GPT-5             & \textbf{92.3} &  7.6 & \textbf{92.3} &  7.6 \\
GPT-5.4-mini      & 76.9 & 11.2 & \textbf{80.8} & 10.6 \\
GPT-5.4           & \textbf{94.2} &  6.8 & 90.4 &  8.2 \\
GPT-5.6-Luna      & 82.7 & 10.2 & \textbf{92.3} &  7.6 \\
GPT-5.6-Sol       & \textbf{96.2} &  6.0 & 90.4 &  8.2 \\
GPT-5.6-Terra     & \textbf{98.1} &  4.9 & 96.2 &  6.0 \\
\bottomrule
\end{tabular}
\caption{Chaining ablation accuracy (\%, $n=52$, chain lengths 2--20).
JSON = JSON tool calling; PTC = programmatic tool calling.
$\pm$ columns show 95\% Wilson CI half-widths. Bold indicates the higher
value per row (ties bolded on both).
$^\dagger$GPT-4.1 PTC collapse caused by \texttt{\textbackslash n} encoding
failure; see text.}
\label{tab:chaining}
\end{table}

\subsection{Parallelism Ablation}
\label{sec:parallelism}

\begin{table}[t]
\centering
\footnotesize
\setlength{\tabcolsep}{3pt}
\begin{tabular}{lrrrr}
\toprule
\textbf{Model} & \multicolumn{2}{c}{\textbf{JSON}} & \multicolumn{2}{c}{\textbf{PTC}} \\
\cmidrule(lr){2-3}\cmidrule(lr){4-5}
 & Acc & $\pm$ & Acc & $\pm$ \\
\midrule
\multicolumn{5}{l}{\textit{Anthropic}} \\
Claude Haiku~4.5  & \textbf{100.0} & 5.5 & \textbf{100.0} & 5.5 \\
Claude Sonnet~4.5 & \textbf{100.0} & 5.5 & \textbf{100.0} & 5.5 \\
Claude Sonnet~4.6 & \textbf{100.0} & 5.5 & \textbf{100.0} & 5.5 \\
Claude Opus~4.8   & \textbf{100.0} & 5.5 & \textbf{100.0} & 5.5 \\
Claude Sonnet~5   & \textbf{100.0} & 5.5 & \textbf{100.0} & 5.5 \\
\midrule
\multicolumn{5}{l}{\textit{OpenAI}} \\
GPT-4o            & \textbf{100.0} & 5.5  & \textbf{100.0} & 5.5 \\
GPT-4.1           &  \textbf{96.9} & 7.5  &  90.6 & 10.5 \\
GPT-5-nano        &  78.1 & 14.0 & \textbf{87.5} & 11.5 \\
GPT-5             &  71.9 & 14.5 & \textbf{96.9} &  7.5 \\
GPT-5.4-mini      & \textbf{100.0} & 5.5  & \textbf{100.0} & 5.5 \\
GPT-5.4           & \textbf{100.0} & 5.5  & \textbf{100.0} & 5.5 \\
GPT-5.6-Luna      & \textbf{100.0} & 5.5  & \textbf{100.0} & 5.5 \\
GPT-5.6-Sol       &  96.9 & 7.5  & \textbf{100.0} & 5.5 \\
GPT-5.6-Terra     & \textbf{100.0} & 5.5  & \textbf{100.0} & 5.5 \\
\bottomrule
\end{tabular}
\caption{Parallelism ablation accuracy (\%, $n=32$). JSON = JSON tool calling;
PTC = programmatic tool calling. Enumeration and overall
accuracy are identical for all models here; aggregation accuracy is discussed
separately in the text. $\pm$ columns show 95\% Wilson CI half-widths. Bold
indicates the higher value per row (ties bolded on both).}
\label{tab:parallelism}
\end{table}

\Codemode{} matches or exceeds baseline accuracy for 13 of \nmodels{} models
on the parallelism ablation ($n = 32$ entries, fan-out counts of 7 to 48 (7, 9, 11, 13, 15, 20, 30, 48)).
Table~\ref{tab:parallelism} separates enumeration accuracy (did the model issue
all $N$ required calls?) from aggregation accuracy (did the model produce the
correct aggregate answer?).

The largest \codemode{} gain is GPT-5, which improves from 71.9\% to 96.9\%.
For this model, baseline accuracy degrades as fan-out count increases above 13,
while \codemode{} maintains near-perfect enumeration across all fan-out levels.
The single model that falls below baseline under \codemode{} is GPT-4.1 (90.6\% vs.
96.9\% baseline), again attributable to the \texttt{\textbackslash n} encoding
failure on parallel scripts that require \texttt{asyncio.gather}.

We observe that \codemode{} enumeration accuracy is high while aggregation accuracy
is lower and noisier across models.
Inspection of outputs shows that models frequently produce the correct aggregation
answer from parametric world knowledge (stating, for example, the top-3
countries by population directly) without executing the enumeration calls.
This is a correct answer by the scorer's standard, but it does not test whether
the model actually executed the tools.
We report both metrics separately and treat enumeration accuracy as the
primary measure of paradigm fidelity.

Token costs cross over at $N \approx 26$. Below this threshold \codemode{}
is more expensive due to its fixed system-prompt overhead. Above it,
\jsonmode{} exceeds \codemode{} as the response must enumerate all $N$
tool-call objects. At $N = 30$, \jsonmode{} uses 3{,}559 tokens versus
3{,}380 for \codemode{}. At $N = 48$, the gap widens to 5{,}097 versus
3{,}535.

\subsection{Context Rot Ablation}

\begin{table}[t]
\centering
\footnotesize
\setlength{\tabcolsep}{2pt}
\begin{tabular}{lrrrrrr}
\toprule
\textbf{Model} & \multicolumn{3}{c}{\textbf{Filtered}} & \multicolumn{3}{c}{\textbf{Flood}} \\
\cmidrule(lr){2-4}\cmidrule(lr){5-7}
 & JSON & PTC & Arg & JSON & PTC & Arg \\
\midrule
\multicolumn{7}{l}{\textit{Anthropic}} \\
Haiku~4.5  & \textbf{93.5} & 83.9 & 48.4 & \textbf{87.1} & 77.4 & 25.8 \\
Sonnet~4.5 & \textbf{90.3} & 83.9 & 58.1 & \textbf{93.5} & 80.6 & 29.0 \\
Sonnet~4.6 & \textbf{93.5} & 83.9 & 83.9 & \textbf{87.1} & \textbf{87.1} & 32.3 \\
Opus~4.8   & 87.1 & 83.9 & \textbf{90.3} & \textbf{83.9} & 80.6 & 41.9 \\
Sonnet~5   & 87.1 & \textbf{87.1} & 80.6 & \textbf{90.3} & 87.1 & 45.2 \\
\midrule
\multicolumn{7}{l}{\textit{OpenAI}} \\
GPT-4o       & \textbf{96.8} & 51.6 & 61.3 & \textbf{90.3} & 74.2 & 16.1 \\
GPT-4.1      & \textbf{83.9} & 51.6 & 41.9 & \textbf{80.6} & 64.5 & 19.4 \\
GPT-5-nano   & \textbf{74.2} & 54.8 & 25.8 & 64.5 & \textbf{71.0} & 16.1 \\
GPT-5        & \textbf{77.4} & 67.7 & 48.4 & 74.2 & \textbf{77.4} & 32.3 \\
GPT-5.4-mini & \textbf{80.6} & 45.2 & 54.8 & \textbf{87.1} & 80.6 & 12.9 \\
GPT-5.4      & 80.6 & \textbf{83.9} & 64.5 & 77.4 & \textbf{77.4} &  9.7 \\
GPT-5.6-Luna & \textbf{83.9} & 77.4 & 35.5 & \textbf{80.6} & 74.2 &  6.5 \\
GPT-5.6-Sol  & 77.4 & \textbf{80.6} & 32.3 & 71.0 & \textbf{80.6} & 12.9 \\
GPT-5.6-Terra & 64.5 & \textbf{80.6} & 35.5 & 71.0 & \textbf{80.6} & 12.9 \\
\midrule
\textbf{Mean $\Delta$} & & & & $-$2.3 & $+$5.5 & $-$32.0 \\
\bottomrule
\end{tabular}
\caption{Context rot ablation accuracy (\%, $n=31$ per condition). Filtered =
entry-relevant schemas only; Flood = 128 total schemas (relevant + corpus decoys).
JSON = JSON tool calling; PTC = programmatic tool calling; Arg = filesystem-discovery.
Bold indicates the highest value among the three paradigms within each condition.
Anthropic model names omit ``Claude'' for space. Mean $\Delta$ = average change from
filtered to flood across all 14 models. 95\% Wilson confidence intervals range
from $\pm$7.8\% to $\pm$16.6\% ($n=31$; widest near 50\% accuracy).}
\label{tab:flood}
\end{table}

The context rot ablation ($n = 31$ entries per condition) compares each
paradigm under two context loads, \textit{filtered} (only the entry's relevant
function schemas) and \textit{flood} (128 total schemas, comprising the relevant
functions plus corpus decoys from unrelated domains).
Table~\ref{tab:flood} reports accuracy under both conditions.

JSON tool calling and \codemode{} are both stable under flooding. The mean accuracy change
from filtered to flood is an absolute $-$2.3\% for JSON tool calling and $+$5.5\% for
\codemode{} (individual model changes range from $-$6.5\% to $+$35.5\% absolute).
The \codemode{} improvement under flood is driven by models that struggled with the
strict type constraints of the filtered condition but found the flood condition's
richer context helped them identify the correct function. This pattern is most
visible in GPT-4.1 ($+$12.9\% absolute), GPT-4o ($+$22.6\% absolute), and GPT-5.4-mini
($+$35.5\% absolute).

For reference, a filesystem-based internal condition degraded by an absolute 32.0\% on
average under flooding, where every model declined, consistent with context navigation
being a structural weakness of file-based tool delivery~\citep{sen2026grep}.

\section{Analysis}

\subsection{Model Generation Predicts \Codemode{} Viability}

The clearest pattern in the results is that \codemode{} accuracy relative to
JSON tool calling tracks model generation, not model family.
All five Anthropic models match or exceed baseline across all three ablation studies,
and so do the three newest GPT-5.6 variants.
The models that fall below baseline under \codemode{} (GPT-4o, GPT-4.1, and
GPT-5.4-mini) share a specific failure. They produce Python code with
\texttt{\textbackslash n} character sequences in place of real newlines, causing
the subprocess to raise a syntax error on any multiline script.

The same system prompt, with the same newline instructions, produces correct
multiline code from GPT-5-nano (a smaller and earlier model) while failing for
GPT-5.4-mini (a larger and later model from a different training run), ruling out
prompt configuration as the cause.
We interpret this as a capability gap, though the specific cause is outside the scope of this paper.

\begin{figure*}[t]
  \centering
  \begin{subfigure}[t]{0.49\linewidth}
    \includegraphics[width=\linewidth]{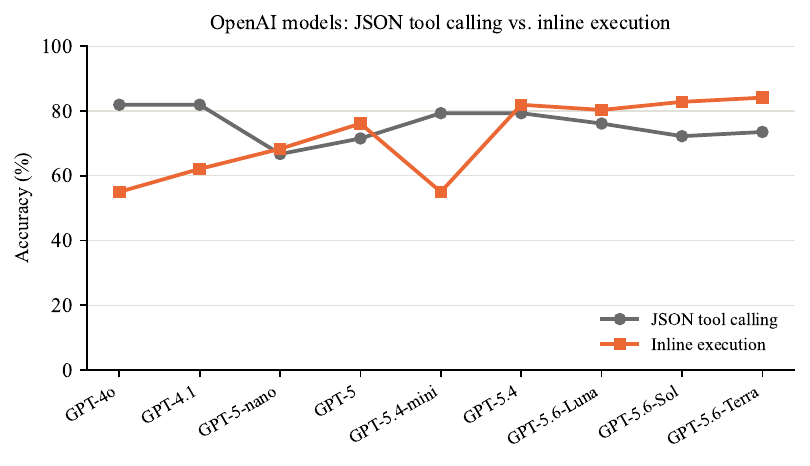}
    \caption{OpenAI models. PTC starts 26.9\% below \jsonmode{} for
      GPT-4o and first crosses baseline at GPT-5-nano before converging fully by the GPT-5.6 generation.}
    \label{fig:openai_line}
  \end{subfigure}
  \hfill
  \begin{subfigure}[t]{0.49\linewidth}
    \includegraphics[width=\linewidth]{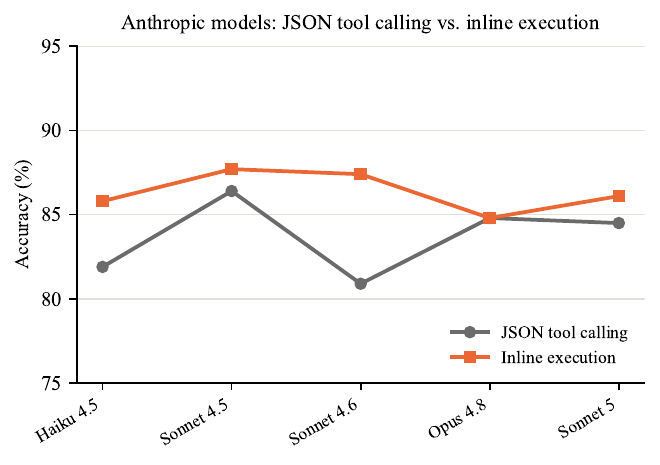}
    \caption{Anthropic models. PTC matches or exceeds \jsonmode{} across
      all five models, with no generation-level gap.}
    \label{fig:anthropic_line}
  \end{subfigure}
  \caption{Accuracy (\%) on the BFCL v4 subset for \jsonmode{} and programmatic tool calling (PTC) by
    model generation. The two families exhibit opposite patterns: OpenAI models
    required several generations to close the gap, while Anthropic models show
    consistent PTC parity from the outset.}
  \label{fig:generation_lines}
\end{figure*}

Figure~\ref{fig:generation_lines} illustrates both trends side by side.

\subsection{\Codemode{} Handles Fan-Out Structurally; JSON Tool Calling Does Not}

\Codemode{} achieves near-ceiling accuracy for GPT-5 on the parallelism
ablation, improving from 71.9\% to 96.9\% while JSON tool calling accuracy degrades
above fan-out counts of 13.
In JSON tool calling, the model must emit a parallel tool-call block in a single
response, and at high fan-out it begins omitting calls.
In \codemode{}, fan-out is expressed as a loop or a sequence of function calls
in Python, which imposes no structural limit on count and benefits from the
model's code-generation training.

The enumeration vs.\ aggregation split reveals a \codemode{} behavior that
departs from the intended design. Several models produce the correct
aggregation answer by drawing on parametric world knowledge rather than
executing the enumeration calls.
We report enumeration accuracy as the primary metric precisely because it
measures whether the model actually invoked the tools, not just whether the
final answer was correct.

To quantify the fan-out count at which this structural advantage emerges for Anthropic models,
we probe Claude Sonnet~5 baseline at $N \in \{60, 70, 72, 75, 100\}$.
Enumeration accuracy is 100\% at $N \leq 70$, drops to 75\% at $N = 72$, and reaches 0\% at
$N = 100$, placing the degradation onset between $N = 70$ and $N = 72$.
\Codemode{} maintains 100\% enumeration accuracy at both $N = 72$ and $N = 100$.
This asymmetry does not appear in GPT-5.6-Sol, which holds 100\% baseline enumeration accuracy
through $N = 100$, suggesting the structural limit is specific to how Anthropic models
serialize parallel tool-call blocks rather than a universal property of JSON tool calling.

\subsection{\Codemode{} Reduces Latency on Chaining Tasks}

\Codemode{} completes chaining entries in roughly half the wall-clock time
of baseline for 13 of \nmodels{} models, with per-entry latency ratios ranging
from 0.32 to 0.96 of baseline.
JSON tool calling requires two inference turns for a sequential chain (one to call
$f_1$, one to receive its result and call $f_2$), while \codemode{} resolves both
calls in a single inference turn followed by one subprocess execution.
GPT-5 is the exception, running at 2.8$\times$ baseline latency under
\codemode{}. Its extended reasoning output inflates generation time enough
to erase the turn-reduction benefit.

\section{Conclusion}

As language model agents increasingly rely on tool use to act beyond their training data,
the choice of tool interface (structured JSON calls versus programmatic tool calling)
has practical consequences for chaining, parallelism, and robustness under real-world conditions.
However, whether these advantages hold systematically across model generations and
real-world task conditions on a standardized benchmark remained an open question.
In this work, we presented a systematic evaluation of programmatic tool calling against
native \jsonmode{} across \nmodels{} language models on \bfcl{},
using typed Python stubs as the tool interface.
Programmatic tool calling matches or exceeds native \jsonmode{} in 11 of 14 models,
with the GPT-5.6 family achieving a 10.7\% improvement over the \jsonmode{} baseline.
It matches or outperforms baseline in 13 of 14 models under parallel fan-out and
holds stable under context rot conditions where baseline degrades 2.3\% on average.
These results demonstrate that programmatic tool calling is a viable and reliable
alternative to \jsonmode{}, with the remaining gap correlating with model generation
rather than model family.

\section{Limitations}

Four limitations bound the claims in this paper.
First, \bfcl{} uses echo-return stubs. Each function returns its arguments
verbatim rather than executing real API calls.
This means we measure argument serialization accuracy, not end-to-end
tool-use correctness. Results may not transfer to settings where return
values affect downstream calls.
Second, the ablation entry counts are small ($n = 31$--$52$ per condition),
so individual model results carry wide confidence intervals and should be
read as directional.
Only the aggregate cross-model patterns (11 of 14 on \bfcl{}, consistent
\codemode{} improvement under flood) are large enough to interpret reliably.
Third, a recent audit found 20\% evaluator-human misalignment in \bfcl{}'s
LLM-judge evaluation mode~\citep{bfclaudit2026}. Our deterministic scorer
avoids this path, but the benchmark's ground-truth labels may still contain
noise that our scorer inherits.
Fourth, \codemode{} carries a fixed input-token overhead relative to \jsonmode{}:
the system prompt embeds the full instruction template as prose, whereas
\jsonmode{} passes function schemas through the API \texttt{tools} parameter,
which some providers count separately from conversation input tokens.
On the chaining ablation, \codemode{} uses 1.5$\times$ the input tokens of
\jsonmode{}. This overhead reverses at high fan-out, as discussed in
Section~\ref{sec:parallelism}. Output token counts do not differ across
paradigms.

\bibliography{references}

\appendix

\section{Per-Category Accuracy on \bfcl{}}
\label{app:per_category}

\begin{table}[H]
\centering
\footnotesize
\setlength{\tabcolsep}{3pt}
\begin{tabular}{lrrrrr}
\toprule
\textbf{Category} & \textbf{$n$} & \textbf{JSON} & \textbf{$\pm$} & \textbf{PTC} & \textbf{$\pm$} \\
\midrule
\textit{simple\_python}   &  40 & 86.1 & 10.7 & \textbf{89.5} &  9.7 \\
\textit{multiple}         &  20 & \textbf{95.4} & 11.1 & 93.9 & 11.9 \\
\textit{parallel}         &  40 & \textbf{85.5} & 10.9 & 71.4 & 13.5 \\
\textit{par.\ multiple}   &  40 & \textbf{85.5} & 10.9 & 71.4 & 13.5 \\
\textit{live\_simple}     &  25 & \textbf{76.6} & 15.9 & 76.3 & 15.9 \\
\textit{live\_multiple}   &  96 & 73.1 &  8.7 & \textbf{83.0} &  7.5 \\
\textit{live\_parallel}   &  24 & \textbf{66.1} & 17.7 & 56.8 & 18.4 \\
\textit{live par.\ mult.} &  24 & \textbf{66.1} & 17.7 & 56.8 & 18.4 \\
\midrule
\textbf{Overall}          & 309 & \textbf{78.6} &  4.6 & 77.0 &  4.7 \\
\bottomrule
\end{tabular}
\caption{Mean accuracy (\%) per \bfcl{} category, averaged across all
\nmodels{} models. JSON = JSON tool calling; PTC = programmatic tool calling. Means are macro-averaged
over models. $\pm$ columns show 95\% Wilson CI half-widths.
Bold indicates the higher value per row. The \textit{parallel} and
\textit{par.\ multiple} gaps are largely driven by the three OpenAI models
with \texttt{\textbackslash n} encoding failures.}
\label{tab:per_category}
\end{table}

\section{\Codemode{} Walkthroughs}
\label{app:walkthroughs}

This appendix shows concrete execution traces to illustrate how the \codemode{}
agent loop operates in each evaluation setting.
Each trace is a verbatim excerpt from a recorded trajectory, lightly condensed
(absolute workspace paths are shortened to \texttt{<workspace>/}).

\subsection{BFCL Single-Call (simple\_python)}
\label{app:walk_bfcl}

\textbf{Task} (\textit{simple\_python\_0}):
\emph{``Find the area of a triangle with a base of 10 units and height of 5
units.''}

The stub module provided to the model contains:
{\footnotesize\begin{verbatim}
def calculate_triangle_area(
        base: int, height: int,
        unit: str | None = None) -> dict:
    """Calculate the area of a triangle."""
    return _rpc.call('calculate_triangle_area',
                     base=base, height=height,
                     unit=unit)
\end{verbatim}
}

The agent emits a single tool call:
{\footnotesize\begin{verbatim}
execute_python(code="""
from stubs import calculate_triangle_area
result = calculate_triangle_area(
    base=10, height=5)
print(__import__('json').dumps(result))
""")
\end{verbatim}
}

The subprocess prints:
{\footnotesize\begin{verbatim}
{"calculate_triangle_area":
    {"base": 10, "height": 5}}
[exit code 0]
\end{verbatim}
}

The stop middleware intercepts the next model call and terminates the loop.
The scorer extracts function name and argument mapping from stdout, compares
against ground truth, and records a single LLM turn for the entry.

\subsection{Chaining Ablation}
\label{app:walk_chaining}

\textbf{Task} (\textit{chaining\_0}):
\emph{``Compute the circumference of a circle with radius 7, then use that
circumference as the side length of a square to find the square's perimeter.''}

The two relevant stubs are \texttt{geometry\_circumference(radius)} and
\texttt{geometry\_square\_perimeter(side)}.

\paragraph{\Codemode{}.}
The agent writes a single script that computes the intermediate value in Python
before passing it to the second stub:
{\footnotesize\begin{verbatim}
execute_python(code="""
import json, math
from stubs import (
    geometry_circumference,
    geometry_square_perimeter)

f1 = geometry_circumference(radius=7)
circ = 2 * math.pi * 7  # 43.982...

f2 = geometry_square_perimeter(side=circ)
print(json.dumps(f1))
print(json.dumps(f2))
""")
\end{verbatim}
}

Subprocess output:
{\footnotesize\begin{verbatim}
{"geometry.circumference": {"radius": 7}}
{"geometry.square_perimeter": {"side": 43.98}}
[exit code 0]
\end{verbatim}
}

Both calls are resolved in a single subprocess, completing the entry in a single LLM turn.

\subsection{Parallelism Ablation}
\label{app:walk_parallelism}

\textbf{Task} (\textit{fanout\_population\_7\_a}):
\emph{``Of these 7 countries — Zorbania, Keltross, Miventa, Drulan, Quelfor,
Hartox, and Bivessa — which 3 have the largest population?''}

The available stub is \texttt{country\_info\_population(country: str) -> dict}.

\paragraph{\Codemode{}.}
The agent issues one \texttt{execute\_python} call that fans out all seven
lookups concurrently and ranks the results in the same script:
{\footnotesize\begin{verbatim}
execute_python(code="""
import asyncio, json
from stubs import country_info_population

pop = country_info_population  # alias
async def main():
    cc = ["Zorbania","Keltross","Miventa",
          "Drulan","Quelfor","Hartox","Bivessa"]
    res = await asyncio.gather(*[
        asyncio.to_thread(pop, country=c)
        for c in cc
    ])
    pairs = sorted(
        zip(cc, [r["population"] for r in res]),
        key=lambda x: x[1], reverse=True)
    top3 = [{"country": c, "population": p}
            for c, p in pairs[:3]]
    print(json.dumps({"answer": top3}))

asyncio.run(main())
""")
\end{verbatim}
}

Subprocess output:
{\footnotesize\begin{verbatim}
{"answer": [
  {"country":"Bivessa","population":891234567},
  {"country":"Zorbania","population":847293441},
  {"country":"Miventa","population":523019876}]}
[exit code 0]
\end{verbatim}
}

All seven stubs are invoked in parallel and Python sorts the results in the same script,
completing the entry in a single LLM turn.

\section{System Prompts}
\label{app:prompts}

\paragraph{JSON tool calling system prompt (condensed).}
The JSON tool calling prompt instructs the model to call the provided tool definitions
using the tool-calling API and to emit one JSON tool-call object per
required function invocation, with no additional prose.
Function schemas are passed as the \texttt{tools} parameter of the API request.

\paragraph{\Codemode{} system prompt (condensed).}
The \codemode{} prompt embeds the source of a typed Python stub module and
instructs the model to call \execpy{} exactly once with a
\texttt{python3 -c '...'} command that imports the module, calls the required
functions with the correct arguments, and prints results as JSON.
The prompt specifies that real newlines (not \texttt{\textbackslash n} escape
sequences) must be used in multiline scripts.
The stop middleware intercepts the first model call after the \execpy{} tool
returns and terminates the agent loop without issuing a second inference.

\paragraph{Stub module design.}
Each stub function is generated from the benchmark's function schema.
It accepts the schema's typed keyword arguments, immediately returns them as a
structured dict, and prints the result to stdout.
No external service is contacted.
The scorer parses stdout to recover function name and argument mapping, then
normalizes and compares against ground truth.

\end{document}